# Path To Gain Functional Transparency In Artificial Intelligence With Meaningful Explainability

Md. Tanzib Hosain
Computer Science & Engineering
American International University-Bangladesh
Dhaka, Bangladesh
tanjibmahammad@gmail.com

Mehedi Hasan Anik
Computer Science & Engineering
American International University-Bangladesh
Dhaka, Bangladesh
hasananik.mh@gmail.com

Sadman Rafi
Computer Science & Engineering
American International University-Bangladesh
Dhaka, Bangladesh
sadmanrafi449@gmail.com

Rana Tabassum
Computer Science & Engineering
American International University-Bangladesh
Dhaka, Bangladesh
rana.tabassum.10536@gmail.com

Khaleque Insia
Electrical & Electronics Engineering
American International University-Bangladesh
Dhaka, Bangladesh
khalequeinsia@gmail.com

Md. Mehrab Siddiky
Electrical & Electronics Engineering
American International University-Bangladesh
Dhaka, Bangladesh
mehrabopi1015@gmail.com

*Abstract*—*Artificial Intelligence (AI) is rapidly integrating into various aspects of our daily lives, influencing decision-making processes in areas such as targeted advertising and matchmaking algorithms. As AI systems become increasingly sophisticated, ensuring their transparency and explainability becomes crucial. Functional transparency is a fundamental aspect of algorithmic decision-making systems, allowing stakeholders to comprehend the inner workings of these systems and enabling them to evaluate their fairness and accuracy. However, achieving functional transparency poses significant challenges that need to be addressed. In this paper, we propose a design for user-centered compliant-by-design transparency in transparent systems. We emphasize that the development of transparent and explainable AI systems is a complex and multidisciplinary endeavor, necessitating collaboration among researchers from diverse fields such as computer science, artificial intelligence, ethics, law, and social science. By providing a comprehensive understanding of the challenges associated with transparency in AI systems and proposing a user-centered design framework, we aim to facilitate the development of AI systems that are accountable, trustworthy, and aligned with societal values.*

## I. INTRODUCTION

The concept of transparency has been a topic of interest, often approached from a technological perspective, in the quest to enhance various aspects of decision-making processes [5]. Previous research has predominantly focused on developing computer-based methods to facilitate transparency. Transparency is seen to empower stakeholders to assess the potential impact of algorithmic decision-making systems and make informed choices about their utilization [3]. However, transparency is often framed in terms of the desired outcomes it aims to achieve, rather than the specific procedures employed to achieve those outcomes.

Algorithmic transparency is essential for stakeholders to comprehend, improve, and question the predictions made by machine learning models. Explainability aims to provide stakeholders with explanations for the actions of such models. However, stakeholders may not be able to determine the accuracy of a model or recognize when it lacks sufficient information for a given task solely by examining its behavior [2]. Explainable machine learning techniques, such as feature significance scores, counterfactual explanations, or analysis of influential training data, can offer insights into model behavior and aid stakeholders in understanding its decision-making process. Nonetheless, the practical implementation of these strategies in businesses remains relatively unexplored [4].

As we move forward, autonomous AI systems are expected to coexist with humans in various domains, providing valuable services. To gain acceptance and trust from users, it is crucial for these systems to be transparent, enabling humans to comprehend their underlying logic. Transparency allows humans to generate meaningful justifications for the decisions and actions of AI systems. Developing algorithms that humans perceive as fair, and embrace is of utmost importance as algorithms increasingly assume greater management and governance responsibilities.

In this paper, our overall contributions are-

1. Firstly, we explain the theoretical concepts related to ensuring transparency. We show the difference of the state f between and in-between. We also outline some traditional problems that arise covering transparency, while blasting the misconceptions of transparency.

2. Secondly, we show different ML factors related to algorithmic transparency, their internal dependencies among each other.

3. Thirdly, we study a case where we showed the dependency between confidence and decision parameter in AI systems while disclosing functional transparency.
4. After that, we show the state of the art of transparency. Finally, provide our design of user-centered compliant-by-design transparency for transparent systems by using empirical approaches to better understand user responses to transparent systems.

## II. EXPLAINABLE ARTIFICIAL INTELLIGENCE

Explainable Artificial Intelligence (XAI) is a set of procedures and techniques that enable human users to understand and trust the output and outcomes of machine learning algorithms [27]. XAI aims to describe an AI model, its anticipated effects, and potential biases in a way that is understandable and transparent to humans. It helps define model correctness, fairness, transparency, and decision-making outcomes supported by AI. By adopting AI explainability, businesses can establish trust and confidence in the development and deployment of AI systems [28].

As AI develops, it becomes challenging for humans to understand and trace the steps taken by algorithms. The calculation process is often treated as a "black box," making it difficult to comprehend [29]. These black box models are generated using data, but even the engineers and data scientists who created the algorithm may not fully understand or describe what happens inside them, including how the AI algorithm reaches specific conclusions.

The goal of an explainable AI (XAI) system is to make its actions more understandable to people by providing justifications. Designing AI systems that are efficient and comprehensible to humans can be guided by the following general principles: the XAI system should be able to describe its capabilities, understandings, and actions, as well as provide insights into the information it acts upon. However, explanations are context-dependent, influenced by the work, user skills, and expectations of the AI system. Thus, interpretability and explainability cannot be defined independently of the domain and are domain specific. Explanations can be complete or partial, with fully interpretable models providing transparent, comprehensive explanations and partially interpretable models shedding light on key aspects of their decision-making process. Interpretable models adhere to "interpretability restrictions" specific to the domain. Examples of partial explanations include variable importance measurements, local models resembling global models in certain instances, and saliency maps.

Understanding how an AI-enabled system produces a specific result offers several benefits. Explainability helps developers ensure that the system operates as intended, may be necessary to meet regulatory standards, and can be crucial in enabling affected individuals to contest or modify decisions [30].

To avoid blindly relying on AI decision-making processes, enterprises must fully comprehend them through model monitoring and accountability. Explainable AI assists humans in better understanding and explaining machine learning (ML), deep learning, and neural networks. ML models are often perceived as opaque "black boxes" that are challenging to understand, particularly when employing neural networks [31]. Addressing bias, often based on race, gender, age, or region, has long been a concern in AI model development. Additionally, AI model performance can drift or deteriorate due to discrepancies between production and training data. Consequently, it is crucial for companies to regularly manage models to enhance AI explainability and assess the impact of deploying such algorithms on their bottom line [26]. Explainable AI also supports end-user trust, model auditability, and effective utilization of AI, while reducing compliance, legal, security, and reputational concerns in production AI. It forms an integral part of responsible AI, providing a framework for the ethical application of AI techniques in real-world businesses with considerations for fairness, model explainability, and accountability [32].

## III. ARTIFICIAL INTELLIGENCE IN BETWEEN WITH OR WITHOUT EXPLANATION

Many companies market simple analytics tools as artificial intelligence, claiming that they can replace human intelligence and provide superior results. However, these tools often involve human labor disguised as artificial intelligence. The use of metaphors and simplifications obscures the fact that human analysts are performing the data analysis behind the scenes. This metaphorical portrayal of mechanical automation may be an inaccurate or exaggerated representation, but it effectively serves as a precise image of the overall concept. This misleading and evolving idea of artificial intelligence forms the basis of marketing gimmicks employed by technology companies. They rely on abstract and oversimplified representations to present a ready-to-use concept of their product, while omitting its true composition and distorting its essence to satisfy the curiosity of end users.

On the other hand, certain systems covertly employ complex data processing and machine learning intentionally. These systems aim to make their influence on decision-making impenetrable, while businesses employ deceptive strategies to contain the action within their own realm. Although these systems could theoretically be designed to be more understandable, the lack of transparency in this era of unprecedented technical complexity cannot be solely attributed to poor communication.

Ultimately, the opacity of machine learning algorithms stems from both institutional self-protection and concealment, as well as the mismatch between the high-dimensional mathematical optimization inherent in machine learning and the requirements of human-scale reasoning and interpretation. The lack of transparency in systems incorporating artificial intelligence solutions is intertwined with the emergence of a new form of automation in cognitive tasks. This fundamental issue goes beyond convoluted storytelling devised by marketing teams or deceptive interfaces and user experience design.

## IV. QUESTIONS ARISE IN EXPLANATION

We have quality assurance and testing techniques, tools, and technologies that can swiftly identify any problems or departures from accepted programming conventions in a normal application development project. We have techniques to continually test our capabilities as we combine them with ever-more complicated systems and application functionality. We can run our apps through regression tests to ensure that new patches and fixes do not cause further problems. However, here is where issues with machine learning models arise. They are not code since we cannot just look at the code to see the faults.

We wouldn't need to train it with data if we were aware of how learning was intended to operate in the first place. We would just start from scratch when coding the model. But that isn't how machine learning models operate. The model's functionality is derived from the data, and we do this by using In a typical application development project, we have established quality assurance and testing techniques, tools, and technologies that can quickly identify any issues or deviations from accepted programming conventions. These methods allow us to continuously test and validate our capabilities as we integrate them into increasingly complex systems and application functionalities. We can run regression tests on our applications to ensure that new patches and fixes do not introduce further problems. However, when it comes to machine learning models, we encounter different challenges.

TABLE I. TRADITIONAL QUESTIONS ON TRANSPARENCY

| Question |
|---|
| What if we are the model's user or consumer? |
| What if the model we are employing isn't working out too well? |
| Was the data used to train it bad? |
| Did the data scientists choose a biased or selected collection of data that doesn't reflect your reality? |
| Should we believe the cloud service provider's model? |
| What about the model that is integrated within the tool we use? |
| What knowledge do we have about the model's construction and iteration process? |
| Do we understand the model's purpose and the use case that the model's creators had in mind? |
| Are we employing the model in the manner that the designers intended? |
| Was a study conducted on the potential effects the model may have on various users? |
| Where did the training data come from? |
| What distinct performance measures are there for various sorts of input data? |
| How does this model fare in real-world testing using different metrics? |
| How can we apply transfer learning to the model? |
| How have measures for model performance evolved over time? |
| Does anybody else use the model? |

Machine learning models are not like traditional code where we can easily identify faults by examining the code itself. Unlike conventional programming, where we start from scratch and write explicit instructions, machine learning models derive their functionality from data. We use algorithms to create the most accurate model possible, given that we need to make generalizations about unseen data. Since we are estimating and generalizing, precision is not guaranteed.

Therefore, fixing bugs alone will not lead us to the perfect model. Instead, we iterate and improve by utilizing better data, fine-tuning hyperparameters, employing advanced algorithms, and increasing computational power. These tools are available to us for improving the model we have created. However, rebuilding the entire model is more challenging and less under our control. Moreover, we may not fully understand why the existing model is not performing well, as it might have been developed with hyperparameter values that worked well for the developers but not for us. As consumers of the model, we have limited options: either use the model as it is or create our own. To establish trust in the models developed by others, there is a need for greater visibility and openness. Model users should carefully consider relying on a model for critical applications due to various concerns regarding model transparency. The lack of answers to these questions is the underlying issue, as there is a lack of openness. Table 1 highlights some of the transparency-related questions that need to be addressed. Addressing these questions will contribute to a more transparent model, addressing real and significant concerns.

V. CONTRASTIVE EXPLANATION: TRANSPARENCY

Transparency can be described in various ways [35]. Related terms such as "explainability" or "XAI," "interpretability," "understandability," and "black box" are sometimes used interchangeably with transparency. Essentially, transparency is a property of an application, indicating the extent to which the inner workings of a system can be theoretically understood. It can also refer to the process of providing user-friendly explanations of algorithmic models and decisions, which contributes to public perception and comprehension of AI. Another perspective on transparency is as a broader socio-technical and normative ideal of "openness" [36].

Transparency refers to the quality of providing access to specific information about how a system operates. However, the relevance of such information from an ethical standpoint depends on the ethical question at hand. Transparency itself is morally neutral; it serves as an ideal. It can take various forms and offer solutions to underlying moral dilemmas.

Some modern machine learning techniques are considered "black box" because users cannot directly observe their inner workings [33]. When these algorithms are used to make decisions that impact individuals, the lack of transparency can be problematic. People have the right to understand how important decisions are made [34]. Consequently, there has been a growing demand for "more transparent AI."

Non-arbitrariness in decision-making is crucial for effective governance in both public and commercial sectors. This principle applies to decisions that significantly affect people in ethical or legal ways. Justifications for why a particular choice was made and the grounds on which it stands indicate non-arbitrariness. Additionally, the ability to challenge and appeal decisions is important, particularly in public governance, as it allows for rectifying any wrongs.

Individuals have the right to access information about the decision-making process to safeguard their agency, freedom, and privacy as per human rights. Freedom encompasses the right to know how one's activities are monitored, what conclusions are being drawn about them, and how those conclusions were reached.

Societal accountability entails managing risks as a collective. There is a moral responsibility to comprehend and anticipate the effects of the technology one develops up to a certain extent. Releasing a harmful system without understanding its potential consequences is not a justifiable course of action. Instead, it is our moral obligation to consider the risks involved.

All these issues can be summarized as requests for sufficient information. Do we know if an algorithmic decision is justifiable, and to what extent? Do individuals understand how assumptions about them are made? To what degree are individuals responsible for the system's actions, and how much knowledge of the system is necessary to assume that responsibility?

The definition of openness or explainability, the level of transparency required for different stakeholders, and other related issues are still topics of debate. The exact definition of "transparency" may vary depending on the context. Whether there are multiple degrees of transparency is still a subject of scientific discussion. Furthermore, the term transparency may be used in different contexts, whether examining the legal implications of unfair biases or discussing characteristics of machine learning systems.

## VI. Problems Arise In Transparency: Why A System Considered As A "Black Box"?

It is quite improbable that we would be able to construct deep learning systems that are entirely transparent because many of the most effective models available now are black box models [38]. As a result, establishing the right amount of openness is the main topic of discussion. Would it be sufficient if algorithms provided the smallest change that could be made to produce a desired result as well as a disclosure of how they arrived at their decision? For instance, if an algorithm rejects someone's request for a social benefit, it should explain why as well as what the person can do to appeal the decision. Transparency also serves a variety of additional purposes in current discussions of machine learning models [39].

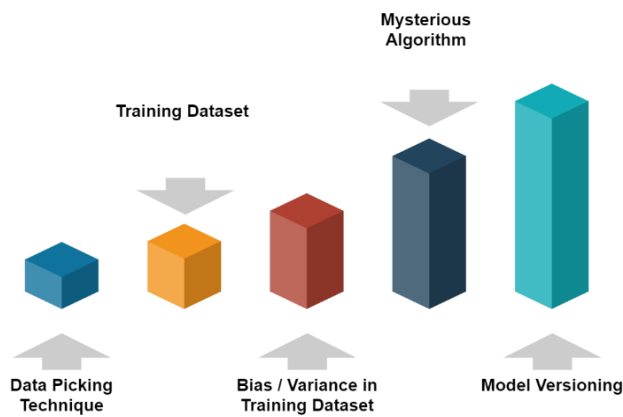

FIG. I. Lackings of Producing a Shadowy Model

It may be crucial for creating legislation or maintaining public confidence in AI. The idea of transparency in AI has given a larger connotation in terms of "comprehensibility" to address these problems. For an algorithm to be comprehensible, or intelligible, it must be possible to explain to individuals who may be impacted by the model how an AI model came to a certain conclusion [40]. One should be able to clearly understand the reasoning behind a decision made using inputs. Fig. 1 shows some possible lackings producing a non-transparent model. After that, the answer for what reason a system is considered a "black-box".

### A. Complexity

The activity of a neural network is stored in dozens or even millions of numerical coefficients in modern AI systems. In most cases, the system picks up these data during the training phase. Even when all the parameters are known, it is almost hard to grasp how the neural network operates since its performance depends on the intricate relationships between these variables.

### B. Developing Difficulties

Even though the AI models that are currently being employed allow explainability to some extent. It could be challenging to design an interface that allows users to receive thorough yet simple explanations.

### C. Risk Concerns

If an attacker carefully constructs an input that breaks the system, they can trick many AI systems. In a system with great transparency, it could be simpler to manipulate the system to produce odd or undesirable consequences. As a result, systems are occasionally purposefully created as black boxes [37].

However, converting notions from algorithms into concepts that people can grasp is notoriously challenging. Legislators in certain nations have debated whether public agencies should reveal the programming codes for the algorithms they employ for automated decision-making. However, most individuals are unable to understand programming codes. Thus, it is difficult to see how publishing codes would increase transparency. Would publishing precise algorithms be more beneficial? Even when the actual techniques are published, there is often little openness, particularly if you do not have access to the model's data.

Today, cognitive and computer scientists create explanations of how and why programs act that are understandable to humans [41]. The creation of tools for data visualization, interactive user interfaces, vocal explanations, or meta-level descriptions of model aspects are a few examples of approaches. These technologies can be of great assistance in increasing the usability of AI applications. There is, however, still a tremendous ton of work being done. This is further complicated by the fact that comprehensibility is reliant on subject- and culture-dependent factors [42]. For instance, diverse cultures use different reasoning when interpreting visuals or drawing conclusions from them. Therefore, it is important for tech developers to have a thorough comprehension of the visual language they employ. Additionally, a lot depends on one's level of algorithmic or data literacy, such one's familiarity with modern technology. The terms used in modern technology are more common in certain cultures than others, and they may be utterly alien to others. There is obviously a need for considerable educational initiatives to improve algorithmic literacy, such as those on "computational thinking," to boost understandability. This user literacy will directly impact how transparent AI systems are in terms of the common users' fundamental knowledge of them. For many people, it could really be the most effective and useful technique to lighten the color of the boxes.

VII. PRECONDITION OF ALGORITHMIC TRANSPARENCY: DATA TRANSPARENCY

Machine learning is opaque, and this applies to both supervised and unsupervised models. Algorithms may significantly increase the efficiency of current business and governance procedures. Every day, algorithms get more intelligent and train themselves to decide quickly and effectively without consulting people.

TABLE II. ANSWER TO SOME TRADITIONAL MYTHS ON ALGORITHMIC TRANSPARENCY

| Myth | Disclosure |
|---|---|
| As organizations are capable of self-regulation, Artificial Intelligence transparency is not required. | Transparency in Artificial Intelligence promotes uniform and predictable regulation. |
| Artificial Intelligence is a machine that lacks emotion, but humans are driven by emotion. | It is possible to encode emotions into an Artificial Intelligence system. |
| That model can't be prejudiced if protected class data isn't used in its construction. | Practitioners of Machine Learning can better understand where to identify biases by having access to protected class data. |
| Humans are known to mess up and make mistakes, but Artificial Intelligence is a computer that can repeat endlessly and consistently. | In any logical sense, Artificial Intelligence is not a machine that can repeat actions endlessly and consistently. |
| Artificial Intelligence transparency puts intellectual property at risk of theft. | Intellectual property disclosure is not necessary for transparency. |
| If Machine Leaning models are seen to act unfairly or biasedly, we will immediately lose the trust of our customers. | Building trust with customers and the public through responsible AI practices. |
| Humans are prejudiced and discriminating, but Artificial Intelligence is a machine that is impartial and neutral. | Computational biases can be introduced into Machine Learning etc. and cause it to behave discriminatorily. |
| While humans have ulterior motives, Artificial Intelligence is a computer devoid of motivation. | Artificial Intelligence has a secret objective that will exploit their human-fueled motivating desires. |
| Humans are capricious and prone to whims, but Artificial Intelligence is a computer that is reliable and will carry out instructions. | Artificial Intelligence is a computer that is not always reliable. |

Information (Data) transparency is "the certainty that data being provided are true and are coming from the official source" as well as "the capability to freely access and operate with data regardless of where they are located or what application developed them". Data transparency gathers reliable information from various sources and presents it in an understandable and practical style [47]. It provides you with a bird's-eye view of the situation. Nothing stays hidden in the dark.

The technological barriers to data sharing are vanishing with time, however this is offset by the growth in the amount of data being produced [48]. Researchers are no longer required to send CD copies of the material they are sharing by mail. Instead, data may be transferred with a single mouse click, in real time, and at a rate of speed that, if it was not so ordinary, would be astounding. In Table 2. clarifications of misconceptions of known algorithmic transparency are shown.

Technology is not a barrier when it comes to medical data [49] [50]. It is important to respect privacy and confidentiality concerns, which are valid concerns [51]. Understanding that behind the data are actual individuals who have the potential to be seriously hurt if their data are mistreated is a key component of Patterns' ethos [52].

Because of this, sensitive data must be subject to protections and limitations, and only those with a "need to know" should have access to it [53].

The use of closed and proprietary databases has been advocated for and the foundation of a great deal of worthwhile scientific study, this is not the problem. Due to selective accessibility, or the fact that professional peer reviewers were given access to the data to verify it, the results based on closed data are still valid. In these situations, the reviewers take on the role of the community, offering quality control checks and reassurances that the data support the results.

Nobody is an expert in every field. Data gathering, annotation, and validation are intricate and time-consuming processes that frequently go unrecognized by the systems in place for academic acknowledgment [54]. To make the usage and creation of research materials like datasets and code more visible and clearer and now the journals now give the option to write descriptor articles on them.

The community that is most likely to utilize the data, including those who are most likely to be peer reviewing it, must be able to reuse it. Creating documentation, annotating data, identifying, and addressing issues with a dataset—all these tasks require time and effort but are required and significant [55].

Data scientists have the expertise and aptitude to swiftly evaluate data and detect the outliers that can suggest issues with the dataset, so researchers serving as peer reviewers don't have to conduct the quality control effort of examining the data in isolation [56]. Collaboration between researchers and data scientists is necessary [57]. An illustration of how a domain expert and a data scientist might collaborate to validate data is as follows: A data scientist would be able to find the negative numbers in a colossally long time series, but the domain expert would know that negative numbers in this stream are a sign that the data are corrupt because negative numbers for this quantity are physically impossible.

Data is better understood when common standards are used [58]. Data is made more useable through common tools and services [59]. By making these things open, others may build on them without having to reinvent the wheel, by modularly linking data and services.

## VIII. ALGORITHMIC TRANSPARENCY

The idea behind algorithmic transparency is that individuals who use, oversee, and are impacted by the systems that use these algorithms should be able to see the elements that drive those decisions. Although the phrase was first used in 2016 to discuss how algorithms are used to determine the content of digital journalism services, the basic idea dates to the 1970s and the emergence of automated systems for calculating consumer credit. In significant ways, algorithms are supplanting or enhancing human decision-making. The use of algorithms to propose anything from goods to buy to music to listen to social network connections has become commonplace. Algorithms, however, are used to make important choices about people's lives, including who receives loans, whose resumes are assessed by humans for potential employment, and the length of prison sentences. Although speed, economy, and even fairness might be advantages of algorithmic decision making, it is a frequent fallacy that algorithms always provide objective judgments. In fact, opaque algorithms have the potential to incorrectly restrict freedom and limit opportunities as well as services [25].

More focus is being paid to algorithmic transparency as significant tasks and procedures are entrusted in algorithmic decision-making systems. Greater openness is recommended by researchers and policymakers as a means of recognizing and preventing several potential harmful consequences of these systems. Experiencing non-obvious knowledge about how and why a system functions as it does and what this means for the system's outputs information that is challenging for a person to understand or personally experience—is a requirement for transparency [3]. Transparency methods offer users the chance to become familiar with features of a system that are often hidden, which has the potential to alter people's perceptions of the system and their interactions with it [43].

### A. Algorithmic Transparency Mechanism

There are numerous various kinds of strategies that have been found to increase algorithmic decision-making transparency. Through regular use of a system, users may become aware of an algorithm [44]. Users occasionally come across surprising or perplexing data that deviates from expectations and suggests algorithmic bias [45]. In other cases, users are encouraged to learn more about the computational outputs so they can develop workarounds to try to prevent undesirable results. Such "organic" awareness, meanwhile, is not systematic nor evenly distributed across users. Algorithm audits, which look at how an algorithmic decision-making system functions and its effects, are another sort of transparency tool propose several levels, each of which might provide a different amount of responsibility and visibility for algorithm audits. However, as system providers frequently exclude the use of auditing techniques in their terms of service, audits must typically be conducted without their assistance. Some claim that platforms deliberately hide information about how they function to protect themselves from rivals or others who try to "game" the system [46].

There are some dependencies for making algorithms transparent, precise, and useful. Precise data, learning about computational output and awareness of the data are the significant factors which can control or manipulate the outcome of an algorithm. Those factors can control undesirable result percentage, expected outcome, algorithmic biasing (Algorithm is biased somehow by the internal or external factors massively).

Mathematically, if the precise data, knowledge on computational output, awareness are proportional to prevent the undesirable data and possible expected outcome, but inversely proportional to the algorithmic biasing.

Table 3 shows Algorithmic decision-making accuracy relationship based on external dependency factors according to the percentage. (↑) Means the percentage increment and (↓) means the decrement of the percentage of input and output factors. So, it can be easily clarified that precise data, knowledge on computational outcome and flawless data entry can enhance the accuracy of the expected outcome, prevent undesired data, and reduce algorithmic biasness.

TABLE III. ALGORITHMIC DECISION-MAKING DEPENDENCIES

| Input \ Output | Preventing Non-Desirable Output | Expecting Output | Biasing Algorithm |
|---|---|---|---|
| Precising Data (↑) | (↑) | (↑) | (↓) |
| Learning about Computational Output (↑) | (↑) | (↑) | (↓) |
| Awareness of Data Entry (↑) | (↑) | (↑) | (↓) |

### B. Factors of Algorithmic Transparency Mechanism

Making a system knowable or transparent is referred to as algorithmic transparency often [60]. In this conception, user behavior or system governance changes are brought about via a transparent method or process [61]. Transparency is, however, occasionally viewed as a condition that results from a process [62]. For instance, as a method of increasing system transparency, the Association for Computing Machinery has included justifications in its list of "Principles for Algorithmic Transparency and Accountability."

Whether "transparency" refers to the method or the result, the cause, or the impact, might be unclear in the literature. In this essay, transparency is the mechanism, and the consequences of transparency are defined in terms of the many sorts of tasks that transparency methods are supposed to be able to do [63]. This enables us to start figuring out what information and justifications explanations may offer to influence specific knowledge and attitudes about systems that use algorithmic decision-making.

Transparency serves the fundamental purpose of making it visible that choices are being made by an algorithm, which raises awareness that interactions with the system are mediated by an algorithm. By explaining what the system is doing, users may notice and understand parts of its behavior that might not be obvious or observable [63]. Users who may not be aware of an algorithm's operations can be especially well-informed by a description alerting them to it.

Transparency techniques also aid users in understanding how the system operates so they may assess the accuracy of the outputs they encounter and recognize inaccurate outputs [3]. Correctness assessments are a result of transparency in that a person cannot independently assess "whether a system is performing as intended and what modifications are required" without a mechanism that can assist an understanding of how the inputs create the outputs. Users should be able to grasp the system's intended outputs and spot flaws or omissions with the aid of a description of how it operates [64].

Transparency can enable judgements about the outputs' sense and indicate that the system's behavior is neither random nor arbitrary in addition to supporting judgments about accuracy [3]. Users are more at ease acting on the outputs when they can understand why a system behaves the way it does and can assess if the system is behaving in accordance with those reasons [65]. This makes the behavior of the system understandable. Users would be better able to comprehend the system's behavior based on perceiving the "truth and intentions" or reasons behind the system's activities, and to recognize when the system is not operating in support of those motives, with the aid of an explanation that supports interpretability [66].

The aim of regulating a system via accountability is emphasized in most of the literature on transparency [67]. Mechanisms for transparency might imply iterative control or individual users believing they are accountable for the outputs of the algorithm [3]. To be effective, a system must be directly answerable to consumers, a justification would present details that support their belief and comprehension that they have direct control over the system's outputs. Ideally, Users are also able to recognize biases because of transparency methods that might have unfavorable effects and empower users can criticize the system and ask questions, presenting evidence for requesting correction [68]. The purpose of this experiment was to determine how user views in an algorithmic decision-making system were affected by four different explanation types (What, How, Why, and Objective), as assessed by the tasks that transparency mechanisms carry out (awareness, correctness, interpretability, accountability). Each transparency function represents a qualitatively distinct view of the system, and certain transparency functions may be more helpful than others in reducing possible adverse impacts of algorithmic decision-making. This experiment is a crucial first step in figuring out how various kinds of information about a system could cause changes in certain user views about transparency.

For advertisements, suggestions, and judgments nowadays, many firms employ algorithms and personal data. Some people are worried that this usage violates people's privacy and endangers both individuals and society. Many people have responded by calling for increased algorithmic openness, or for businesses to be more transparent and open about how they utilize algorithms and personal data [60].

IX. FACTORS OF EXPLANATION-PROBABILITY WITH RISK

In Machine Learning, we use the term uncertainty to describe our ignorance about a particular result of interest. To understand and measure uncertainty, we employ probability-based reasoning techniques. Probability is seen by the Bayesian school of thinking as individual levels of conviction that a desired result will materialize [16] [2].

Fig. 2 and Fig. 3 show our result of probabilistic confidence and decision graph of while tackling risk. It shows how confidence and decision are co-related with each other while working on transparency risk related to some intelligent system. The probabilistic confidence which we have used as $C$ and the observed data as D. The probabilistic confidence function $C(D)$ can be expressed using Bayesian probability as follows:

$$C(D) = P(Model \mid D) \qquad (1)$$

Here, $P(Model \mid D)$ is the posterior probability of the model given the observed data $D$. It represents the confidence in the model's parameters (weights) after considering the data. On the other hand, the decision as $Dec$ and the model parameters (weights) as $W$. The decision function $Dec(W)$ can be expressed as a threshold-based decision using Bayesian probability as follows:

$$Dec(W) = \{1, if\ P\ (Class=1 \mid W) > Threshold\ 0,\ if\ P\ (Class=1 \mid W) <= Threshold\} \qquad (2)$$

Here, $P(Class=1 \mid W)$ is the probability of the positive class (class 1) given the model parameters $W$. Threshold is a predefined value used to make the binary decision. If the probability of the positive class is higher than the threshold, the decision is $1$ (positive), otherwise, it is $0$ (negative). As we work for this taking observation of some traditional algorithm in Artificial Intelligence transparency; it is a big concern that the two function shows that the confidence vs decision is inversely proportional. We have got the relationship between functions as follows:

$$Dec(D) = k / C(D) \qquad (3)$$

Here, $k$ is a constant representing the strength of the inverse relationship. The constant $k$ can be adjusted to control the strength of the inverse relationship between confidence and decision. A higher value of $k$ will result in a stronger inverse relationship, while a lower value of $k$ will result in a weaker inverse relationship. Bayesian approaches clearly specify a hypothesis space of plausible models a prior then apply deductive logic to update these priors given the actual data. This is often accomplished in parametric models, such as Bayesian Neural Networks [2] [20] [21], by considering model weights as random variables rather than single values and assigning them a prior distribution. The conditional probability informs us how well each weight setting explains our observations given certain observed data. The prior is updated using the likelihood, which results in the posterior distribution over the weights. Through the process of marginalization, a prediction for a test point is created. For each configuration of weights, the prediction is weighed according to the posterior density of that weight. Model uncertainty arises from the discrepancy between predictions made using several reasonable weight choices. Both epistemic and aleatoric uncertainty are captured by the predicted posterior distribution. Due to their adaptability and scalability to massive volumes of data, Neural Network have recently gained respect among the Machine Learning community. According to frequentists, probabilities represent how frequently we would see the result if we repeated our observation repeatedly [17] [2] [18]. For the benefit of end users, uncertainty resulting from frequentist and Bayesian approaches conveys identical information in practice [2] [19] and is frequently addressed equally in subsequent tasks. Research communities and application areas utilize different measures to convey uncertainty. Although a comprehensive predictive distribution has a lot of information, it may not always be desired. To provide information concerning uncertainty, summary statistics of the predictive distribution are frequently utilized. The class probabilities make up the predictive distribution for classification. These innately convey our level of confidence as a result. Predictive entropy, on the other hand, separates our predictions from their uncertainty and only informs us of the latter. A predicted mean and error bar is frequently used to summarize the predictive distribution for regression. These frequently represent the predictive distribution's standard deviation or a few percentiles.

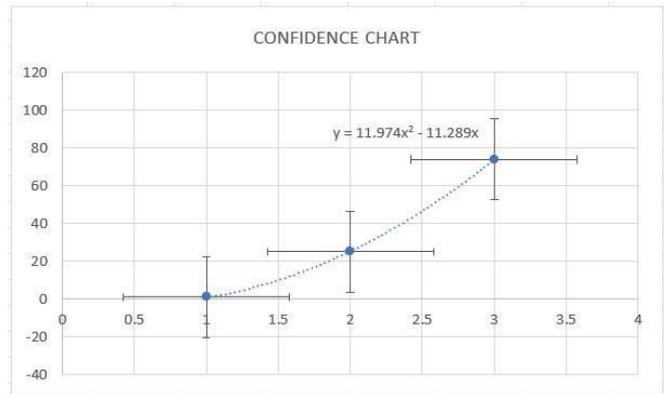

FIG. 2. Probabilistic Confidence Graph

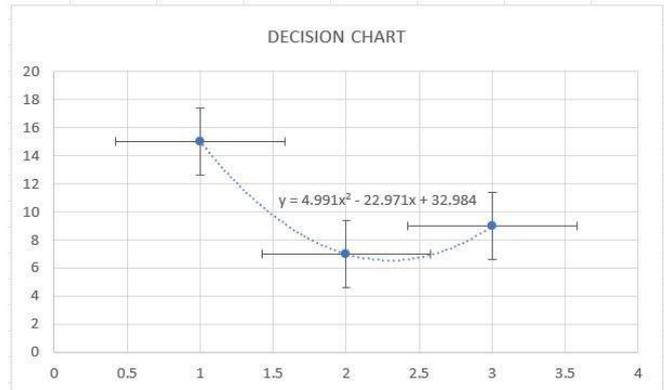

FIG. 3. Probabilistic Decision Graph

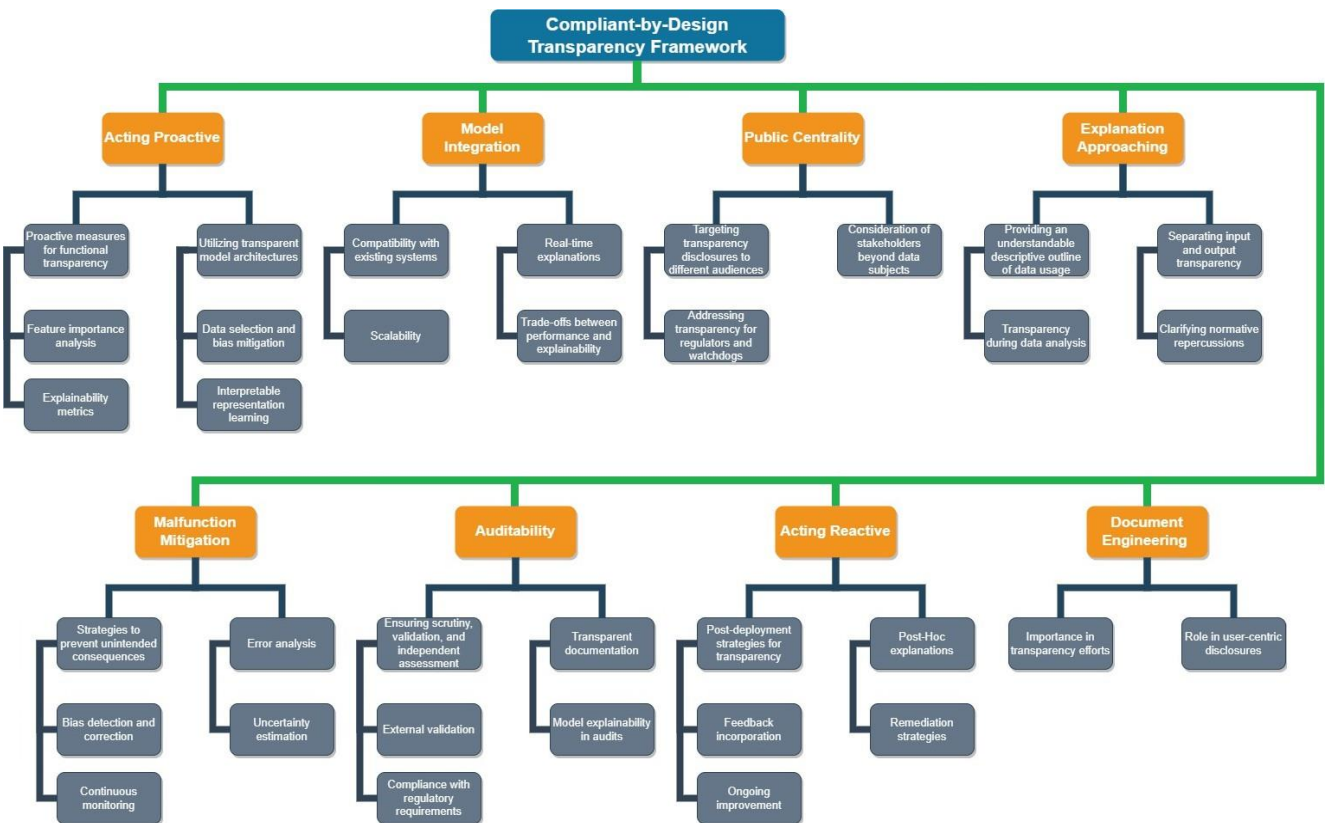

FIG. 4. Functional Transparency Model

## X. TOWARD STATE OF THE ART OF FUNCTIONAL TRANSPARENCY

Organizations have long struggled with the transparency dilemma in several contexts, including security and privacy. The inner workings of artificial intelligence models should be more transparent, have been argued by academics and practitioners in recent years, and for many excellent reasons. Fairness, prejudice, and trust are concerns that may all be lessened through transparency. These issues have all gotten more attention recently. It's a common belief in the field of data analytics that more data is always better. However, data itself is frequently a source of liability in risk management. The same is starting to be true with artificial intelligence. Organizations have long struggled with the transparency dilemma in several contexts, including security and privacy. All they must do is modernize their processes for AI. Companies aiming to use artificial intelligence need to understand that openness comes at a price. This is not to indicate that transparency is not desirable; rather, it is to point out that it also has drawbacks that must be properly appreciated. These expenses ought to be considered by a larger risk model that determines how to interact with explainable models and how much of the model's details are made public. Additionally, businesses need to understand that security is a growing risk in the AI industry. As my colleagues and I recently discussed at the Future of Privacy Forum, as AI becomes more extensively used, more security flaws and bugs will undoubtedly be found. In fact, one of the largest long-term obstacles to the adoption of AI may be security [24].

## XI. COMPLIANT-BY-DESIGN ADHERING FUNCTIONAL TRANSPARENCY

Calling for AI system transparency alone has little practical benefit. Transparency standards must be turned into actionable activities. To offer such useful advice, the compliant-by-design transparency principles were created. As a result, our approach concentrates on basic design criteria, user-oriented system information providing, and the organizational management of system transparency. The Compliant-by-Design Openness Framework's ultimate objective is to promote the advantages of transparency while simultaneously minimizing its drawbacks. When including the transparency requirement in the creation of AI systems, system designers, particularly engineers, and organizational stakeholders are held to these standards. Design-side requirements line up with other stakeholders' rights to openness and inspection, particularly users and third parties. The transparency principles' primary intended audience is engineers. Fig. 4 shows our design model of functional transparency.

### A. Acting Proactive

Acting proactive refers to the proactive measures taken to ensure functional transparency and meaningful explainability in artificial intelligence (AI) systems. This subsection focuses on preemptive strategies that can be employed during the design, development, and deployment stages of AI models. The aim is to integrate interpretability and transparency features from the inception of the AI system, rather than trying to retrofit explanations after the model has been deployed. Key components of acting proactive include:

1. Transparent Model Architectures: Utilizing model architectures that inherently promote interpretability, such as decision trees, rule-based models, or explainable neural networks.

2. Feature Importance Analysis: Identifying and analyzing the most influential features in the model's decision-making process, providing insights into its behavior.

3. Data Selection and Bias Mitigation: Ensuring that the training data used for the AI model is diverse, representative, and free from biases that could lead to unfair or unethical decisions.

4. Explainability Metrics: Defining quantitative metrics to assess the level of explainability of the AI system and continuously improving it based on these metrics.

5. Interpretable Representation Learning: Exploring techniques that learn feature representations that are easier to understand and interpret.

### B. Model Integration

Model integration refers to the process of seamlessly incorporating explainable AI models into real-world applications and systems. This addresses the challenges and opportunities of deploying interpretable AI models in complex environments where transparency is crucial. Key considerations in model integration include:

1. Compatibility with Existing Systems: Ensuring that the explainable AI models can be easily integrated with existing infrastructures, APIs, and frameworks without causing significant disruptions.

2. Real-time Explanations: Enabling the AI system to provide meaningful explanations in real-time to users, facilitating better trust and understanding.

3. Scalability: Designing interpretable models that can scale to handle large volumes of data and users without compromising on their transparency.

4. Trade-offs: Addressing trade-offs between model performance and explainability, as more complex models may offer higher accuracy but lower interpretability.

### C. Public Centrality

The relational character of transparency communications must be respected by AI system developers. Therefore, it is not enough to simply provide information; one must also consider who is most likely to receive and understand it. The transparency obligation will change based on who the anticipated information recipient is, as is noted. Although the primary recipients of personal information processing are the individuals whose data is being used, it is important to be aware that other stakeholders must also be considered. These stakeholders may include distinct categories of system users, regulators, watchdogs, or the public, depending on the nature and function of the system in question. The information offered to the affected individual, where specific information needs for decision-making on the individual will be the key focus, will need to differ from information intended at the wider public, where queries about general functioning would be prevalent. Unlike the public, government agencies, regulators, and independent watchdogs would receive information that was not only specific to their needs but also possibly more technical in nature. This is because it can be assumed that these institutions, unlike the public, have relevant specialized expertise at their disposal. If transparency disclosures are targeted at regulators or trained third parties that represent the public interest rather than data subjects themselves, they may have a greater impact. The need to make decisions understandable to affected people, the public, regulatory bodies, or watchdogs is also closely related to avoiding public mistrust and suspicion.

*D. Explanation Approaching*

This principle calls for the provision of an understandable descriptive outline of the data being used by the system and the ways in which it is being used, including information on what stages of data processing are inspectable and where human discretion, intervention, or oversight occurs in the system. This is necessary considering the potential technical limitations on the explainability of decision-making of complex AI systems. The system can use data for several purposes.

The system's operation and any biases are defined by the previous stages, even if the stakeholders' main transparency objectives pertain to the consumption stage. Therefore, one could argue that to fully comply with the transparency requirements at this time, access should be given to the working protocol analysts use for these preliminary parts of the prediction tasks, such as the display of the data that was utilized for analysis. Transparency in input and output can be separated. Input openness often comes at an excessive cost and offers little benefits, in part because of the potentially enormous volume. Technical transparency during data analysis might imply that the program used to make automated choices should be made public. If specialized software is employed, this gets increasingly difficult since requiring such openness can violate people's legal rights to their intellectual property. Systems with characteristics that may be examined by relevant parties without disclosing the code or input data can help with these problems. This might aid in the design of systems that are open about the characteristics that matter for a certain automated judgment while safeguarding personal information and keeping trade secrets hidden. A need for transparency might be the revelation of the real methods and procedures used to use the data. The precision that the Fairness, Accountability and Transparency in Machine Learning [22] principles define should be part of this. Depending on whether such openness is intended for professionals or laypeople whose information demands are more general and less technical, the criteria for delivering such information may seem different. Making information about a system intelligible for the broader public is regarded to be the definition of explainability. Where there is human judgment, control, or supervision, transparency information must also be disclosed.

The practice of explainability to ensure transparency begins through properly explaining the process of processing data with meaningful insights. The data processing process is iterative and may involve multiple rounds of analysis and refinement to improve the quality of insights obtained from the data. Additionally, data processing can also include data cleaning and preprocessing steps to ensure the data's accuracy, completeness, and consistency before performing analysis and extraction.

1. Data Collection: Data collection is the first step in the process of data processing. In this step, raw data is gathered from various sources, such as databases, sensors, logs, surveys, social media, or any other relevant data sources. The collected data can be structured, semi-structured, or unstructured.

2. Data Aggregation: Once the data is collected, the next step is data aggregation. In this step, data is organized and combined to form a larger dataset. Aggregation may involve summarizing or grouping the data based on specific attributes or criteria. The purpose of aggregation is to reduce data volume and to provide a more concise and manageable dataset for analysis.

3. Data Analysis: After data aggregation, the data is analyzed to derive meaningful insights and draw conclusions. Data analysis involves various techniques and methods, such as statistical analysis, machine learning, data mining, or any other domain-specific analysis approaches. The goal is to uncover patterns, trends, correlations, or anomalies within the data.

4. Data Extraction: Data extraction is the final step in the process. In this step, the relevant insights and information obtained from data analysis are extracted and presented in a usable format. This can involve generating reports, visualizations, or data summaries that provide actionable insights to support decision-making processes.

Transparency is required when it comes to the influence of human decision-making, which may be prone to biases, such as when it comes to the development and selection of training data. A right to be exempt from automated decision-making that might have a significant impact on one's life legally or in other ways is specifically included in the General Data Protection Regulation [23]. Therefore, disclosures about when and how such human interaction occurs must be included in transparency standards. Along with disclosure of the descriptive aspects of data processing, the variables that influence decision-making should be considered. These normative norms carry implicit meanings in them. Clarifying normative repercussions is necessary for the implicit decision-making criterion to be transparent. They discuss the limitations of various explanation approaches before settling on an interpretation of explanation that is intricately linked to the idea of interpersonal justifiability rather than aiming for a precise descriptive representation. The following crucial aspects of justifiability must be addressed why certain data are a normatively acceptable basis for drawing inferences and why these inferences are normatively acceptable and relevant for the chosen explanation. When evaluating data processing, it is important to look both ahead, at the data sources that are connected to the desired outputs, and backward, at the underlying assumptions that influence how we, as data subjects, are seen and assessed by outside parties. It is important to closely consider how data sources and conclusions relate to one another. The level of privacy invasion caused by processing, the counter intuitiveness of the inferences, the specific intentions driving the processing, the use of potentially discriminatory features, the potential repercussions of deriving sensitive information from non-sensitive information, the acceptability of doing so according to norms, and the racial and ethnic composition of the source data could all be relevant issues to discuss in the context of justification.

*E. Malfunction Mitigation*

Malfunction mitigation deals with strategies to prevent or minimize unintended and harmful consequences of AI systems due to misunderstandings, biases, or errors in the decision-making process. This subsection emphasizes the need to identify and address potential issues before they lead to negative outcomes. Important components of malfunction mitigation include:

1. Error Analysis: Conducting thorough error analysis to understand the types of mistakes the AI model is prone to make and developing targeted solutions to rectify them.

2. Bias Detection and Correction: Implementing mechanisms to detect and correct biases in the model's predictions, ensuring fairness and ethical decision-making.

3. Uncertainty Estimation: Incorporating uncertainty estimation techniques to communicate the confidence levels of the AI system's predictions, especially in critical scenarios.

4. Continuous Monitoring: Setting up continuous monitoring and auditing processes to identify and rectify malfunctioning AI models in real-time.

*F. Auditability*

Auditability focuses on ensuring that AI systems are subject to scrutiny, validation, and independent assessment of their performance and behavior. This subsection highlights the importance of third-party audits and evaluations to gain trust and confidence in AI applications. Key aspects of auditability include:

1. Transparent Documentation: Providing comprehensive documentation of the AI model, including its architecture, training data, hyperparameters, and explainability mechanisms.

2. External Validation: Allowing external auditors and evaluators to assess the AI system's performance and transparency, promoting accountability and reducing bias.

3. Model Explainability in Audits: Ensuring that explanations provided by the AI model are understandable and meaningful for auditors to assess the system's decision-making process.

4. Compliance and Regulatory Requirements: Aligning with relevant regulations and guidelines for AI transparency, explainability, and auditability.

*G. Acting Reactive*

Acting reactive refers to measures taken in response to unforeseen issues, ethical concerns, or legal requirements regarding AI transparency and explainability. This subsection deals with post-deployment strategies to address challenges and continuously improve AI systems' transparency. Key components of acting reactive include:

1. Post-Hoc Explanations: Developing techniques to explain AI model decisions retrospectively, even for black-box models, to understand the factors influencing predictions.

2. Feedback Incorporation: Integrating user feedback and input into the AI model to improve its transparency and correct potential misconceptions or biases.

3. Remediation Strategies: Implementing strategies to rectify issues related to transparency and explainability discovered after the AI system is deployed in real-world scenarios.

4. Ongoing Improvement: Establishing a feedback loop for continuous improvement of the AI model's transparency and explainability over time.

*H. Document Engineering*

Both academia and business are paying more and more attention to the problem of transparency in machine learning models and datasets [9] [15]. The objective has frequently been to increase visibility into ML models and datasets through source code disclosure [15] [10], contribution histories [11] [15], the introduction of ML-driven data analysis techniques, and the introduction of varied supervision [15] [12]. In terms of regulation from government entities throughout the world, transparency and explainability of model outputs via the lens of datasets has become a major problem. However, in research and industrial contexts, efforts to develop standardized, real-world, and long-lasting methods for transparency that provide value at scale encounter little success. This reflects the limitations imposed by the diversity of aims, processes, and backgrounds of the various stakeholders involved in the dataset and artificial intelligence system life cycles [1] [13] [14].

Engineering must also pick up new abilities. The ability to style papers according to specifications in the forms needed by the contract, professional society, or internal norms is now just as important as the ability to write effectively and concisely. Engineering owes readers an additional duty. It must recognize that it is an essential part of the new cost-competitive emphasis that is changing the way engineering companies conduct business: If it doesn't increase the bottom line, change it so that it does, or get rid of it. It must also understand the standard writing skills of audience identification and writing clearly to those audiences.

The process of defining, creating, and implementing the information models that support document-centric applications is known as document engineering [6] [7]. To create documents that are more generic and resilient, it is necessary to describe various information and data sources using new document models. A key paradigm in the shift from people-oriented documents to technical schemas, document engineering is supposed to help make information flows more uniform, organized, compatible, and visible across stakeholders.

To simplify and automate document sharing inside and across organizations, Document Engineering looks for commonality in forms and transaction documentation. Relationships inside organizations are thought of as a series of document exchanges where the parties are aware of one another's papers. Documents are seen in this sense as disclosing the inputs and outputs of business operations, where they act as the organization's public face [6] [7]. Document Engineering technique is based on the needs of the stakeholders and involves the breakdown of current information sources into the reconstruction of associated documents.

Because Document Engineering seeks to make document interchange more organized, consistent, and intelligible, it stands to reason that the technique may be used to produce disclosures that are user focused. Since disclosures are essentially documents, we may examine instances of present disclosure procedures and then reconstruct them as user-facing documents with interfaces that are more logical and better suit the needs of their receivers. Therefore, after the content of those papers is defined, Document Engineering works in conjunction with design techniques, documentation approaches, or any number of other viable methods to develop and adapt the look of those disclosures. although the usage context varies.

Once more, Document Engineering is only one method for producing disclosures that are more relevant. Our goal in describing how we may use this technique is not to make the case that this is how disclosures need to be made or that Document Engineering will be appropriate in every circumstance. Instead, we highlight one-way interfaces offer a fresh approach to information disclosures that are tailored to receivers' requirements.

TABLE IV. Process Of Document Engineering

| Organization Level | Analyzing Business Process | Analyzing Context of Service | |
|---|---|---|---|
| Process Level | Applying Pattern on Process Models | Implementing Process and Documents Models | |
| | Assembling Document Models | | |
| Information Level | Assembling Document Components | Analyzing Document Components | Analyzing Documents |
| | Conceptual Models | Physical Models | Implementation Models |

1. Process of Document Engineering: Understanding multi-modal datasets that provide models nuance rapidly becomes more difficult as research and business progress toward large-scale models capable of many downstream functions. The responsible and informed deployment of models, especially those in human-facing contexts and high-risk domains, requires a clear and full knowledge of a dataset's origins, development, intent, ethical issues, and evolution. However, the documentation's clarity, succinctness, and thoroughness frequently bear the weight of this knowledge. All the dataset's documentation must be consistent and comparable, and as a result, documentation must be seen as a standalone user-centric product. The Document Engineering process, which is crucial because disclosure interfaces require an iterative approach to be refined and adjusted as systems, usage contexts, and requirements change and as the overall landscape of transparency regimes continue to change. This is like how the applications themselves are developed. In Table 4, we can see that effective documentation can be maintained through its 4x4 matrix requirements.

Processes in document engineering [6] [7], which we now described being applied to a disclosure context.

1. Conceptual Analysis: Analyzing the settings in which the process' output, in this case the disclosure, will be utilized is a component of document engineering. When discussing disclosures, this entails figuring out who the possible stakeholders are and what they would hope to gain from the disclosure, which serves as the basis for the subsequent analysis. It is crucial to consider the legal restrictions that the disclosure must adhere to. The results of it are frequently universally relevant across a variety of disclosures, and as a result, the settings of applications may be comparable across a variety of organizations and industries. As was previously said, a common need for revealed information is that it should be contextually suitable and provide information that is accurate, relevant, proportionate, and understandable [7] [8]. The way disclosure data is presented will inevitably have an influence on each of these areas, and it can help determine how disclosure interfaces might better enable stakeholders to participate in the processing of their personal data. Afterward, investigating present business procedures Here, we considered the kinds of data supplied as well as how this data is categorized, structured, and portrayed within existing disclosures using the variety of disclosures we acquired. With the help of these patterns, we were able to begin developing a suggestive model of the kinds of data that our model disclosure should probably include as well as how this data should be arranged and presented to create a more contextually relevant disclosure. Afterward, analyze the documents. This required analyzing the information in the disclosures we acquired, including their structure to determine how relevant material is categorized and whether visualizations or other modalities could be more appropriate given the context of the recipient's needs. To determine where the disclosures may more closely match user experiences.
2. Implementation: Begin developing component prototypes. This entails looking at several presentation options for disclosure data that have been reassembled into more relevant forms. The components are then put together to create a document model, which helps to fully grasp how the final document should be organized and how relevant components should be grouped. The new interface must then be operationalized and put into use. This requires the organization to implement the designed disclosure.

## XII. LIMITATIONS

Every area of research, business, and government is still being transformed by the data revolution. We are becoming more conscious of the need to utilize data and algorithms responsibly in line with laws and ethical standards due to the enormous influence that data-driven technology has on society. There is a significant drive for organizations to increase their accountability and transparency. To this end, several transparency laws compel businesses to provide specific information to interested parties. With the use of this information, organizational activities are meant to be monitored, overseen, scrutinized, and challenged. However, it is important to note that these disclosures are only useful to the extent that their recipients find them valuable. However, the disclosures of tech/data-driven organizations are frequently extremely specialized, scattered, and hence of little use to anybody outside experts. This lessens the impact of a revelation, disempowers people, and thwarts larger transparency goals.

After conducting the research on explainable AI and the transparency of AI some fundamental outcomes have been founded like the importance of the transparency of AI, data and Algorithmic transparency mechanisms and the designs of the transparency techniques. But some major and minor areas are untouched in the research. The limitations of the research are:

1. Cons of the over transparency of AI and algorithm.
2. Transparency controlling mechanisms are also a limitation of the research.
3. Uncertainty accuracy for the algorithmic prediction.
4. Trustworthiness of the automation and advanced algorithms for artificial intelligence.

## XIII. FUTURE RESEARCH OPPORTUNITIES

Though lots of studies are done and new studies are running to provide more transparency in Artificial Intelligent systems, there are still work to be done. Some of the future working sectors for transparency are listed below:

1. Researching how people feel about privacy and openness- How do people see privacy and transparency in regards to learning and control? When do individual actions align or deviate from the declared value?
2. Peer transparency vs hierarchical openness as a method of control- How do peer relationships differ from hierarchical ones in terms of the behavioral impacts of transparency/privacy? Do the outcomes alter depending on whether the witness and the witnessed are part of a stable structure or the general public?
3. Examining the value of trust and openness in society- How do trust and culture affect learning and control when it comes to openness and privacy?
4. Using different degrees of analysis to produce effective methods for balancing transparency and privacy- How does the observation of a person, organization, region, or entire company affect the perceived dangers and advantages of openness vs privacy? Is there a relationship between an institution's fundamental ambidexterity and how privacy is constructed?
5. Observed characteristics and their influence on how transparency and privacy affect behaviors- Do various characteristics affect how transparency and privacy work? Do different nations experience privacy and transparency differently? Do millennials behave differently in terms of openness and privacy?
6. Looking into the causes and processes of behavioral reactions to privacy and transparency- Does privacy serve an instinctive or strategic requirement in humans? How might neuroscience aid our understanding of how people behave when faced with openness or privacy?
7. Researching the effects of different types of transparency- Does the operation of physical such as open offices and digital such as open data transparency differ? How do process transparency and result transparency varies from one another? How do transparencies that are only momentary and persistent function differently? How do delayed transparency and rapid transparency differ from one another? How do mandatory and voluntary transparency differ from one another? How do anonymous and personally identifiable transparency vary from one another?
8. Methodological possibilities and difficulties- Mixed-method studies that blend privacy and transparency research approaches. Real-time field experiments that record information on both observer and observed viewpoints as organizations expand openness.

## XIV. CONCLUSION

In this research paper, we have highlighted the significance of transparency and explainability in artificial intelligence (AI) systems. As AI becomes an integral part of our lives, understanding how these systems make decisions is crucial for ensuring fairness and accuracy. Achieving transparency, however, poses numerous challenges that must be addressed. We have proposed a user-centered, compliant-by-design approach for transparent systems. Our argument is that developing transparent and explainable AI systems is a complex endeavor that necessitates collaboration across multiple disciplines, including computer science, artificial intelligence, ethics, law, and social science. By emphasizing the need for transparency, we underscore the importance of enabling stakeholders to comprehend and trust AI systems. This research contributes to the growing body of knowledge surrounding transparency in AI and provides a foundation for future studies in this area. Moving forward, it is imperative that researchers, policymakers, and industry professionals work together to develop robust frameworks and guidelines for transparency and explainability. Only through interdisciplinary collaboration and ongoing research efforts can we ensure that AI systems are accountable, fair, and trustworthy, promoting their responsible and ethical deployment in society.

## FUNDING

This research did not receive any outside funding or support. The authors report no involvement in the research by the sponsor that could have influenced the outcome of this work.

## AUTHORS` CONTRIBUTIONS

All authors have participated in drafting the manuscript. All authors read and approved the final version of the manuscript. All authors contributed equally to the manuscript and read and approved the final version of the manuscript.

## CONFLICT OF INTEREST

The authors certify that there is no conflict of interest with any financial organization regarding the material discussed in the manuscript.

## ACKNOWLEDGMENT

Thanks to Dr. M. F. Mridha.

## REFERENCES

[1] Ehsan, U., Liao, Q. V., Muller, M., Riedl, M. O., & Weisz, J. D. (2021, May). Expanding explainability: Towards social transparency in ai systems. In Proceedings of the 2021 CHI Conference on Human Factors in Computing Systems (pp. 1-19).

[2] Bhatt, U., Antorán, J., Zhang, Y., Liao, Q. V., Sattigeri, P., Fogliato, R., ... & Xiang, A. (2021, July). Uncertainty as a form of transparency: Measuring, communicating, and using uncertainty. In Proceedings of the 2021 AAAI/ACM Conference on AI, Ethics, and Society (pp. 401-413).

[3] Rader, E., Cotter, K., & Cho, J. (2018, April). Explanations as mechanisms for supporting algorithmic transparency. In Proceedings of the 2018 CHI conference on human factors in computing systems (pp. 1-13).

[4] Bhatt, U., Xiang, A., Sharma, S., Weller, A., Taly, A., Jia, Y., ... & Eckersley, P. (2020, January). Explainable machine learning in deployment. In Proceedings of the 2020 conference on fairness, accountability, and transparency (pp. 648-657).

[5] Springer, A., & Whittaker, S. (2019, March). Progressive disclosure: empirically motivated approaches to designing effective transparency. In Proceedings of the 24th international conference on intelligent user interfaces (pp. 107-120).

[6] Urquhart, C., & Spence, J. (2007). Document Engineering: Analyzing and Designing Documents for Business Informatics and Web Services. Journal of Documentation, 63(2), 288-290.

[7] Norval, C., Cornelius, K., Cobbe, J., & Singh, J. (2022). Disclosure by Design: Designing information disclosures to support meaningful transparency and accountability. In 2022 ACM Conference on Fairness, Accountability, and Transparency (pp. 679-690). ACM.

[8] Marsh, C. H. (1999). The engineer as technical writer and document designer: The new paradigm. ACM SIGDOC Asterisk Journal of Computer Documentation, 23(2), 57-61.

[9] Biasin, E. (2022). ACM Conference on Fairness, Accountability, and Transparency (ACM FAccT): Doctoral Consortium Session.

[10] Antunes, N., Balby, L., Figueiredo, F., Lourenco, N., Meira, W., & Santos, W. (2018). Fairness and transparency of machine learning for trustworthy cloud services. In 2018 48th Annual IEEE/IFIP International Conference on Dependable Systems and Networks Workshops (DSN-W) (pp. 188-193). IEEE.

[11] Barclay, I., Taylor, H., Preece, A., Taylor, I., Verma, D., & de Mel, G. (2021). A framework for fostering transparency in shared artificial intelligence models by increasing visibility of contributions. Concurrency and Computation: Practice and Experience, 33(19), e6129.

[12] Hutchinson, B., Smart, A., Hanna, A., Denton, E., Greer, C., Kjartansson, O., Barnes, P., & Mitchell, M. (2021). Towards accountability for machine learning datasets: Practices from software engineering and infrastructure. In Proceedings of the 2021 ACM Conference on Fairness, Accountability, and Transparency (pp. 560-575). ACM.

[13] Hutchinson, B., Smart, A., Hanna, A., Denton, E., Greer, C., Kjartansson, O., Barnes, P., & Mitchell, M. (2021). Towards accountability for machine learning datasets: Practices from software engineering and infrastructure. In Proceedings of the 2021 ACM Conference on Fairness, Accountability, and Transparency (pp. 560-575). ACM.

[14] Felzmann, H., Fosch-Villaronga, E., Lutz, C., & Tamò-Larrieux, A. (2020). Towards transparency by design for artificial intelligence. Science and Engineering Ethics, 26(6), 3333-3361.

[15] Pushkarna, M., Zaldivar, A., & Kjartansson, O. (2022). Data cards: Purposeful and transparent dataset documentation for responsible AI. In 2022 ACM Conference on Fairness, Accountability, and Transparency (pp. 1776-1826). ACM.

[16] MacKay, D. J. C. (2003). Information theory, inference and learning algorithms. Cambridge University Press.

[17] Bland, J. M., & Altman, D. G. (1998). Bayesians and frequentists. BMJ, 317(7166), 1151-1160.

[18] Pek, J., & Van Zandt, T. (2020). Frequentist and Bayesian approaches to data analysis: Evaluation and estimation. Psychology Learning & Teaching, 19(1), 21-35.

[19] Xie, M., & Singh, K. (2013). Confidence distribution, the frequentist distribution estimator of a parameter: A review. International Statistical Review, 81(1), 3-39.

[20] MacKay, D. J. C. (1992). Bayesian interpolation. Neural Computation, 4(3), 415-447.

[21] Palakkadavath, R., & Srijith, P. K. (2021). Bayesian generative adversarial nets with dropout inference. In Proceedings of the 3rd ACM India Joint International Conference on Data Science & Management of Data (8th ACM IKDD CODS & 26th COMAD) (pp. 92-100).

[22] FAT. (2018). Fairness, accountability, and transparency in machine learning. Retrieved December 24, 2018.

[23] Voigt, P., & Von dem Bussche, A. (2017). The EU general data protection regulation (GDPR): A practical guide (1st Ed.). Springer International Publishing.

[24] Burt, A. (2019). The AI transparency paradox. Harvard Business Review. Retrieved from https://bit.ly/369LKvq

[25] Garfinkel, S., Matthews, J., Shapiro, S. S., & Smith, J. M. (2017). Toward algorithmic transparency and accountability. Communications of the ACM, 60(9), 5-5.

[26] Speith, T. (2022). A review of taxonomies of explainable artificial intelligence (XAI) methods. In 2022 ACM Conference on Fairness, Accountability, and Transparency (pp. 2239-2250). ACM.

[27] Gunning, D., Stefik, M., Choi, J., Miller, T., Stumpf, S., & Yang, G.-Z. (2019). XAI—Explainable artificial intelligence. Science Robotics, 4(37), eaay7120.

[28] Das, A., & Rad, P. (2020). Opportunities and challenges in explainable artificial intelligence (XAI): A survey. arXiv preprint arXiv:2006.11371.

[29] von Eschenbach, W. J. (2021). Transparency and the black box problem: Why we do not trust AI. Philosophy & Technology, 34(4), 1607-1622.

[30] Gade, K., Geyik, S. C., Kenthapadi, K., Mithal, V., & Taly, A. (2019). Explainable AI in industry. In Proceedings of the 25th ACM SIGKDD

International Conference on Knowledge Discovery & Data Mining (pp. 3203-3204).

[31] Nielsen, M. A. (2015). Neural networks and deep learning (Vol. 25). Determination Press.

[32] Wang, Y., Xiong, M., & Olya, H. (2020). Toward an understanding of responsible artificial intelligence practices. In Proceedings of the 53rd Hawaii International Conference on System Sciences (HICSS) (pp. 4962-4971). Hawaii International Conference on System Sciences.

[33] Rai, A. (2020). Explainable AI: From black box to glass box. Journal of the Academy of Marketing Science, 48, 137-141.

[34] Wachter, S., & Mittelstadt, B. (2019). A right to reasonable inferences: Re-thinking data protection law in the age of big data and AI. Columbia Business Law Review, 494.

[35] Ball, C. (2009). What is transparency? Public Integrity, 11(4), 293-308.

[36] Bostrom, N. (2017). Strategic implications of openness in AI development. Global Policy, 8(2), 135-148.

[37] Rosenberg, N. (1982). Inside the black box: Technology and economics. Cambridge University Press.

[38] Rudin, C. (2019). Stop explaining black box machine learning models for high stakes decisions and use interpretable models instead. Nature Machine Intelligence, 1(5), 206-215.

[39] Burrell, J. (2016). How the machine 'thinks': Understanding opacity in machine learning algorithms. Big Data & Society, 3(1), 2053951715622512.

[40] Yampolskiy, R. V. (2020). Unexplainability and incomprehensibility of AI. Journal of Artificial Intelligence and Consciousness, 7(02), 277-291.

[41] Winograd, T., Flores, F., & Flores, F. F. (1986). Understanding computers and cognition: A new foundation for design. Intellect Books.

[42] Chromá, M. (2008). Two approaches to legal translation. In Language, Culture and the Law: The Formulation of Legal Concepts across Systems and Cultures (Vol. 64, pp. 303).

[43] Eslami, M., Rickman, A., Vaccaro, K., Aleyasen, A., Vuong, A., Karahalios, K., Hamilton, K., & Sandvig, C. (2015). "I always assumed that I wasn't really that close to [her]": Reasoning about invisible algorithms in news feeds. In Proceedings of the 33rd Annual ACM Conference on Human Factors in Computing Systems (pp. 153-162).

[44] Goodman, B., & Flaxman, S. (2017). European Union regulations on algorithmic decision-making and a "right to explanation". AI Magazine, 38(3), 50-57.

[45] Dupret, G. E., & Piwowarski, B. (2008). A user browsing model to predict search engine click data from past observations. In Proceedings of the 31st Annual International ACM SIGIR Conference on Research and Development in Information Retrieval (pp. 331-338).

[46] Gillespie, T. (2018). Custodians of the Internet: Platforms, content moderation, and the hidden decisions that shape social media. Yale University Press.

[47] Bhushan, B., Khamparia, A. Sagayam, K. M., Sharma, S. K., Ahad, M. A., & Debnath, N. C. (2020). Blockchain for smart cities: A review of architectures, integration trends and future research directions. Sustainable Cities and Society, 61, 102360.

[48] Fawcett, S. E., Wallin, C., Allred, C., Fawcett, A. M., & Magnan, G. M. (2011). Information technology as an enabler of supply chain collaboration: A dynamic-capabilities perspective. Journal of Supply Chain Management, 47(1), 38-59.

[49] Boonstra, A., & Broekhuis, M. (2010). Barriers to the acceptance of electronic medical records by physicians: From systematic review to taxonomy and interventions. BMC Health Services Research, 10(1), 1-17.

[50] Yarbrough, A. K., & Smith, T. B. (2007). Technology acceptance among physicians: A new take on TAM. Medical Care Research and Review, 64(6), 650-672.

[51] Ulman, Y. I., Cakar, T., & Yildiz, G. (2015). Ethical issues in neuromarketing: "I consume, therefore I am!". Science and Engineering Ethics, 21, 1271-1284.

[52] Watson, L. C. (1976). Understanding a life history as a subjective document: Hermeneutical and phenomenological perspectives. Ethos, 4(1), 95-131.Schwartz, Paul M. "European data protection law and restrictions on international data flows." Iowa L. Rev. 80 (1994): 471.

[53] Schwartz, P. M. (1994). European data protection law and restrictions on international data flows. Iowa L. Rev., 80, 471.

[54] Diaz, O., Kushibar, K., Osuala, R., Linardos, A., Garrucho, L., Igual, L., Radeva, P., Prior, F., Gkontra, P., & Lekadir, K. (2021). Data preparation for artificial intelligence in medical imaging: A comprehensive guide to open-access platforms and tools. Physica Medica, 83, 25-37.

[55] Walsham, G. (2006). Doing interpretive research. European Journal of Information Systems, 15(3), 320-330.

[56] Grus, J. (2019). Data science from scratch: First principles with python. O'Reilly Media.

[57] Wang, D., Weisz, J. D., Muller, M., Ram, P., Geyer, W., Dugan, C., Tausczik, Y., Samulowitz, H., & Gray, A. (2019). Human-AI collaboration in data science: Exploring data scientists' perceptions of automated AI. Proceedings of the ACM on Human-Computer Interaction, 3(CSCW), 1-24.

[58] Kasabov, N. K. (2014). NeuCube: A spiking neural network architecture for mapping, learning and understanding of spatio-temporal brain data. Neural Networks, 52, 62-76.

[59] Abraham, M. J., Murtola, T., Schulz, R., Páll, S., Smith, J. C., Hess, B., & Lindahl, E. (2015). GROMACS: High performance molecular simulations through multi-level parallelism from laptops to supercomputers. SoftwareX, 1, 19-25.

[60] Lepri, B., Oliver, N., Letouzé, E., Pentland, A., & Vinck, P. (2018). Fair, transparent, and accountable algorithmic decision-making processes: The premise, the proposed solutions, and the open challenges. Philosophy & Technology, 31, 611-627.

[61] Chen, I. J., & Popovich, K. (2003). Understanding customer relationship management (CRM): People, process and technology. Business Process Management Journal, 9(5), 672-688.Yang, Jing, and Ava Francesca Battocchio. "Effects of transparent brand communication on perceived brand authenticity and consumer responses." Journal of Product & Brand Management 30, no. 8 (2021): 1176-1193.

[62] Yang, J., & Battocchio, A. F. (2021). Effects of transparent brand communication on perceived brand authenticity and consumer responses. Journal of Product & Brand Management, 30(8), 1176-1193.

[63] Shin, D. (2021). The effects of explainability and causability on perception, trust, and acceptance: Implications for explainable AI. International Journal of Human-Computer Studies, 146, 102551.

[64] Lim, B. Y., Dey, A. K., & Avrahami, D. (2009). Why and why not explanations improve the intelligibility of context-aware intelligent systems. In Proceedings of the SIGCHI Conference on Human Factors in Computing Systems (pp. 2119-2128).

[65] Landwehr, C. E., Bull, A. R., McDermott, J. P., & Choi, W. S. (1994). A taxonomy of computer program security flaws. ACM Computing Surveys (CSUR), 26(3), 211-254.

[66] Wu, L., & Chen, J. L. (2005). An extension of trust and TAM model with TPB in the initial adoption of on-line tax: An empirical study. International Journal of Human-Computer Studies, 62(6), 784-808.

[67] Hess, D. (2007). Social reporting and new governance regulation: The prospects of achieving corporate accountability through transparency. Business Ethics Quarterly, 17(3), 453-476.

[68] Arrieta, A. B., Díaz-Rodríguez, N., Del Ser, J., Bennetot, A., Tabik, S., Barbado, A., García, S., et al. (2020). Explainable Artificial Intelligence (XAI): Concepts, taxonomies, opportunities and challenges toward responsible AI. Information Fusion, 58, 82-115.